\documentclass[pdflatex,sn-mathphys-num]{sn-jnl}

\usepackage{graphicx}%
\usepackage{multirow}%
\usepackage{amsmath,amssymb,amsfonts}%
\usepackage{amsthm}%
\usepackage{mathrsfs}%
\usepackage[title]{appendix}%
\usepackage{xcolor}%
\usepackage{textcomp}%
\usepackage{manyfoot}%
\usepackage{booktabs}%
\usepackage{algorithm}%
\usepackage{algorithmicx}%
\usepackage{algpseudocode}%
\usepackage{listings}%

\begin{document}

\title[Transfer Learning via Auxiliary Labels with Application to Cold-Hardiness Prediction]{Transfer Learning via Auxiliary Labels with Application to Cold-Hardiness Prediction}


\author*[1]{\fnm{Kristen} \sur{Goebel}}\email{goebelk@oregonstate.edu}

\author[2]{\fnm{Paola} \sur{Pesantez-Cabrera}}\email{p.pesantezcabrera@wsu.edu}

\author[3]{\fnm{Markus} \sur{Keller}}\email{mkeller@wsu.edu}

\author[1]{\fnm{Alan} \sur{Fern}}\email{afern@oregonstate.edu}

\affil[1]{\orgdiv{Electrical Engineering and Computer Science}, \orgname{Oregon State Univeristy}, \orgaddress{\city{Corvallis}, \state{Oregon}, \country{United States}}}

\affil[2]{\orgdiv{Electrical Engineering and Computer Science}, \orgname{Washington State University}, \orgaddress{\city{Pullman}, \state{Washington}, \country{United States}}}

\affil[3]{\orgdiv{Viticulture and Enology}, \orgname{Washington State University}, \orgaddress{\city{Pullman}, \state{Washington}, \country{United States}}}

\abstract{Cold temperatures can cause significant frost damage to fruit crops depending on their resilience, or cold hardiness, which changes throughout the dormancy season. This has led to the development of predictive cold-hardiness models, which help farmers decide when to deploy expensive frost-mitigation measures. Unfortunately, cold-hardiness data for model training is only available for some fruit cultivars due to the need for specialized equipment and expertise. Rather, farmers often do have years of phenological data (e.g. date of budbreak) that they regularly collect for their crops. In this work, we introduce a new transfer-learning framework, Transfer via Auxiliary Labels (TAL), that allows farmers to leverage the phenological data to produce more accurate cold-hardiness predictions, even when no cold-hardiness data is available for their specific crop. The framework assumes a set of source tasks (cultivars) where each has associated primary labels (cold hardiness) and auxiliary labels (phenology). However, the target task (new cultivar) is assumed to only have the auxiliary labels. The goal of TAL is to predict primary labels for the target task via transfer from the source tasks. Surprisingly, despite the vast literature on transfer learning, to our knowledge, the TAL formulation has not been previously addressed. Thus, we propose several new TAL approaches based on model selection and averaging that can leverage recent deep multi-task models for cold-hardiness prediction. Our results on real-world cold-hardiness and phenological data for multiple grape cultivars demonstrate that TAL can leverage the phenological data to improve cold-hardiness predictions in the absence of cold-hardiness data.}




\maketitle

\section{Introduction}\label{sec1}

\begin{figure}[t]
  \centering
  \includegraphics[width = 0.9\columnwidth]{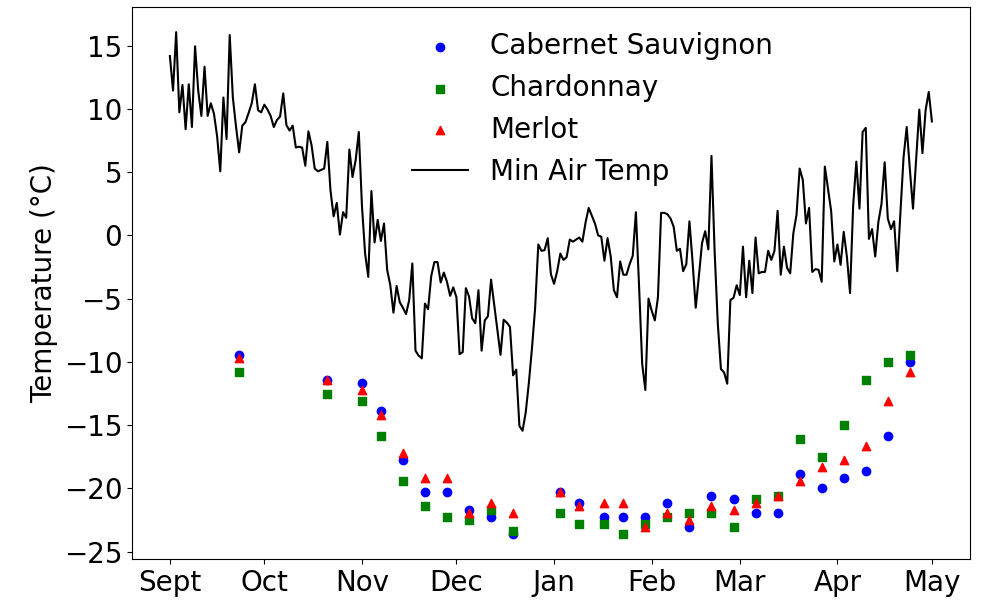}
  \caption{Cold-hardiness data for three grape cultivars compared to minimum air temperature. The data points are $LTE_{50}$, or the temperature at which 50\% of the grape buds will experience lethal freezing (see Section \ref{sec:related}). Note how the cold-hardiness has a general seasonal trend as well as responses to the local weather. Different cultivars show distinct cold-hardiness responses, indicating that growers need models that are tuned for their individual cultivars.
  }
\label{fig:cold-hardiness-example}
\end{figure}

This work is motivated by the objective of assisting farmers in making informed decisions to protect perennial fruit crops, such as grapes, from lethal cold temperatures. A grapevine's resistance to frost damage, or cold hardiness, changes during the dormant season as the plant develops and the weather changes. To avoid frost damage, farmers deploy expensive frost-mitigation methods, such as wind machines and heaters. Ideally, these decisions should be informed by the crop's current cold-hardiness. However, obtaining accurate information about cold hardiness on a specific day is challenging. While the general trend is for cold hardiness to increase in the fall, making plants more resilient, and decreases in the spring, the day-to-day cold hardiness can fluctuate significantly in response to local weather patterns (Figure \ref{fig:cold-hardiness-example}). Further, directly measuring cold hardiness requires expertise and expensive equipment, which is typically only available for research.

Since farmers can't measure cold hardiness, they must rely on personal experience and/or cold-hardiness models to make management decisions. Current state-of-the-art models predict daily cold hardiness based on weather data and take different forms, including scientific models \cite{c:ferguson2014} and deep neural networks \cite{c:saxena2023}. 
Importantly, the cold-hardiness response of plants can vary significantly across different cultivars. Thus, cold-hardiness models are cultivar specific, since they must be trained on ground-truth cold-hardiness data for a particular cultivar.
There are many cultivars for which farmers have no model to benefit from as such training data is unavailable.


The primary goal of our work is to enable cold-hardiness modeling for currently unmodeled cultivars by leveraging auxiliary data that most farmers already collect. In particular, phenological information, such as the date of budbreak, can easily be observed and is commonly recorded in farm logs. When scientists collect cold-hardiness data for selected cultivars, they also record phenological information. This raises the question of whether we can leverage the well-studied cultivars with both cold-hardiness and phenological data in order to predict cold hardiness for cultivars with only phenological data. For this purpose, we introduce a novel transfer learning paradigm, \emph{Transfer via Auxiliary Labels (TAL)}. In our application of TAL, tasks correspond to cultivars and the source tasks have training data for both the primary label (cold hardiness) and auxiliary label (phenology). Given the source tasks, the aim of TAL is to (transfer) learn a model that can predict the primary label for a target task (i.e. new cultivar) when that target task has only auxiliary labels in the training data.  

Despite the vast literature on transfer learning, to our knowledge, the TAL paradigm has not been considered in prior work (see Section \ref{sec:related}). This is particularly surprising, since TAL appears relevant to many analogous problems in agriculture as well as other very different application domains. For example, in a medical setting tasks might correspond to individual patients, primary labels might correspond to data from an expensive test, and auxiliary labels might correspond to data from a related inexpensive device or cheap infrequent test. Given a data set of (source) individuals with both the expensive and inexpensive labels, TAL would allow for transfer to (target) individuals who only have inexpensive data. 

The main contribution of our work is to propose multiple approaches to TAL and explore their effectiveness for cold-hardiness prediction. Our approaches build on recent multi-task learning work, which achieved state-of-the-art performance by jointly learning models over multiple cultivars. We extend that work by including phenological labels during training and using model-selection and model-averaging approaches to predict cold hardiness for the target cultivar using only auxiliary data. Our experiments show that TAL improves predictions on new cultivars compared to natural baselines. 

In summary, this paper makes the following contributions: 
\begin{enumerate}
\item Introduces the novel TAL paradigm for transfer learning. 
\item Develops principled model-selection and model-averaging approaches to TAL along with baselines.
\item Evaluates and compares the TAL approaches on the real-world problem of cold-hardiness prediction for grapes. 
\item Identifies a model-averaging approach that significantly outperforms baselines and will be beta tested by growers on the widely used AgWeatherNet \cite{c:agnet}. 
\end{enumerate}

\section{Background and Related Work}
\label{sec:related}

\subsection{Grape Cold Hardiness Modeling}

The cold hardiness of a plant indicates its resistance to cold damage, which varies throughout dormancy. In grapevines, this damage includes lethal bud freezing, which decreases crop yield. Cold hardiness is difficult to measure, scientists must use methods like differential thermal analysis (DTA) \cite{c:mills2006}, which requires expensive and complicated equipment. Bud samples are taken from the vines and placed in a controlled refrigerator where the temperature is lowered to identify the temperatures at which 10\%, 50\%, and 90\% of the buds freeze, denoted $LTE_{10}$, $LTE_{50}$, and $LTE_{90}$,  respectively. Figure \ref{fig:cold-hardiness-example} shows the $LTE_{50}$ data for three cultivars that was collected at approximately bi-weekly intervals throughout a single season. 

Scientists have developed predictive models for cold hardiness (e.g. \cite{ferguson_dynamic_2011,c:ferguson2014}), which integrate plant biology concepts to find a relation between daily temperatures and changes in cold hardiness. These models have tunable cultivar-specific parameters, which are optimized to match the cultivar's cold-hardiness data. Recently, this data has been used to train neural networks for cultivar-specific cold-hardiness prediction \cite{c:saxena2023}, which allows for the incorporation of weather variables beyond temperature (e.g. humidity, wind speed, etc.). These models have shown higher average accuracy compared to the prior scientific models, however, all of these models are limited to cultivars for which enough DTA data has been collected. 

Since many cultivars do not have associated DTA datasets, our work considers how to leverage phenological data for cold-hardiness prediction. During the dormancy period, grapevine development is marked by several phenological events, or biological milestones, which are influenced by the climate. These events are easily observable and often recorded by farmers. In this paper, we consider four events: first swell, full swell, budbreak, and first leaf. Scientific models similar to those for cold hardiness have been developed for budbreak prediction \cite{c:ferguson2014}, which is a critical event to predict since the plants are particularly vulnerable to cold at that stage. Recently, neural networks have been trained for budbreak prediction \cite{saxena2023b}, which is treated as binary classification of whether budbreak has occurred on or before each day. However, those efforts did not consider leveraging auxiliary phenological data for cold-hardiness prediction, which is the primary goal of our work.


\subsection{Learning with Auxiliary Data}

Since our framework is based on using auxiliary labels, we review the most closely related uses of auxiliary data from prior work.  Auxiliary data has been used in various ways to improve the performance of a primary learning task in many applications, including computer vision \cite{zhang2014facial}, natural language processing \cite{mikolov2013distributed}, and reinforcement learning \cite{jaderberg2016reinforcement}. However, this prior work has primarily focused on single-task learning scenarios without an emphasis on transfer.  Perhaps the closest prior work on using auxiliary data is multi-domain matrix completion, where sparse primary labels are estimated with the assistance of paired auxiliary labels (e.g. \cite{pan2011transfer}). While this problem is sometimes referred to as transfer from auxiliary to primary labels, it is more similar to data imputation than to the transfer-learning setting we consider in this paper. In particular, the approach is highly specialized for imputing the missing values in a fixed primary matrix, rather than transfer learning of complex functions as considered in our work.  



\subsection{Related Transfer Learning Approaches}

There are a wide variety of transfer learning paradigms \cite{pan2009survey}, though we have not yet found prior work that is applicable to the paradigm introduced in this paper. Our TAL paradigm has some similarity to transductive transfer learning. In both paradigms the primary labels are available for the source task(s) and are not available in the target tasks. However, transductive transfer learning does not leverage auxiliary labels for transfer, but rather focuses on adjusting for differences between the source and target input data distributions. Some example of transductive approaches include: weighting the source data or models \cite{Sun2011}, using source models to label target data \cite{tan2013multi}, estimating source data similarity \cite{aljundi2017expert}, and training the source models to reach consensus on the target data \cite{zhuang2009cross}. All of these methods are based on the idea that the input data distributions for the source and target differ and do not use auxiliary data, making them inapplicable to our problem. 

Another related transfer formulation is learning from multiple sources, e.g. \cite{crammer2008}, where the source and target problems share the same data types for inputs and output labels, but the underlying function between the inputs and outputs for each task differ. Solutions aim to identify the set of source tasks that are most similar to the target task in order to leverage that source data for the target. While the different source tasks in this formulation could be considered as a form of auxiliary data for learning the target, this notion of auxiliary data is very different from that in TAL. In particular, in TAL the target only has auxiliary labels, which may have a data type that differs from the primary labels. It could be interesting to significantly modify this work to address the TAL problem, but it is not straightforward and beyond the scope of this paper.

\section{TAL Problem Formulation}

\begin{figure}[t]
  \centering
  \includegraphics[width = 0.9\columnwidth]{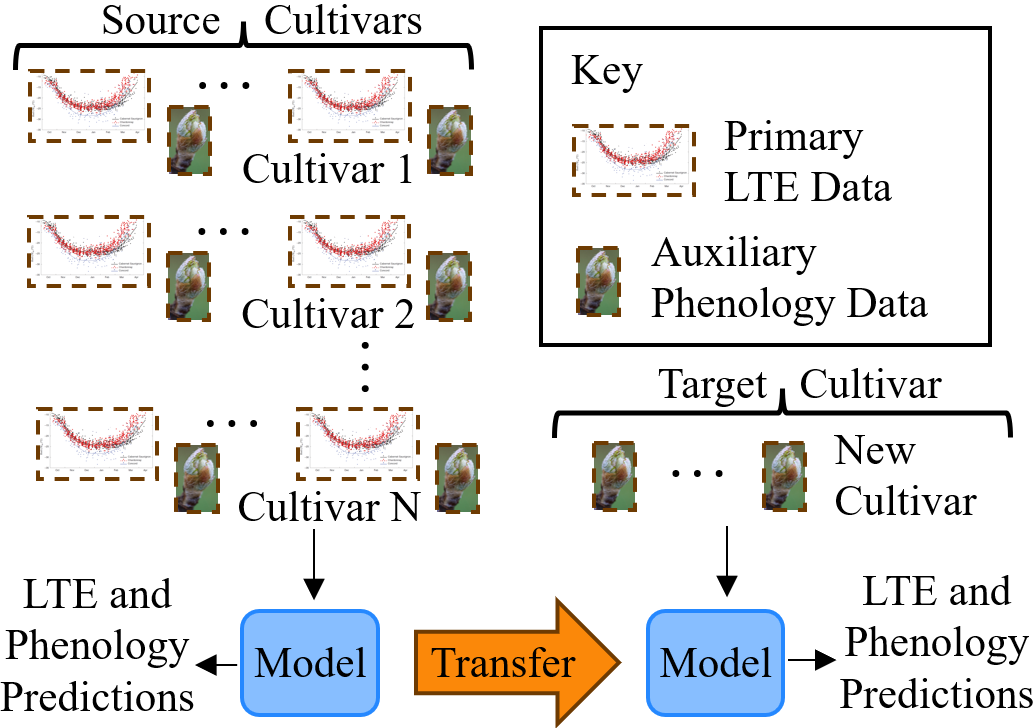}
  \caption{Overview of TAL. We learn a model for the source cultivars using the corresponding primary LTE and auxiliary phenology data. The new target cultivar only has phenology data, so we cannot directly learn a target LTE model and instead transfer knowledge from the source cultivar models.}
\label{fig:TAL}
\end{figure}

In this section we formally specify the TAL problem, which is instantiated in Section \ref{sec:tal-for-cold-hardiness} for our motivating problem of cold-hardiness prediction.  At a high level, the TAL problem assumes a source set of prediction tasks whose training data includes paired primary and auxiliary labels. The goal is to transfer from the source tasks to make primary label predictions for a new target task that has only auxiliary labels in its training set. Figure \ref{fig:TAL} illustrates this process.


Formally, we let $T$ denote the set of all possible tasks, indexed by $i$. Each task corresponds to an unknown ground truth function $g_i : \mathcal{X} \rightarrow \mathcal{Y}\times \mathcal{Z}$ that maps a shared input space $\mathcal{X}$ to a pair of outputs, one from the primary output space $\mathcal{Y}$ and one from the auxiliary output space $\mathcal{Z}$. Typically, the primary outputs are intended for direct use in the end application, while the auxiliary outputs are intended as a type of side information to help in learning primary outputs. In our transfer setting, the set of source tasks is denoted by $T_s \subset T$. Each source task $i\in T_s$ has a corresponding training dataset $D_i = \{(x,y,z)_{k} \in \mathcal{X}\times\mathcal{Y}\times \mathcal{Z} | k \in {1,...,N_i}\}$, where the inputs $x$ are drawn from an unknown distribution and the outputs $y$ and $z$ are governed by the unknown function $g_i$. 

The TAL formulation aims to transfer from the source tasks to a new target $i^* \notin T_s$, which comes with a training data set $D_{i^*} = \{(x,z)_{k} \in \mathcal{X}\times\mathcal{Z} | k \in {1,...,N_{i^*}}\}$. The distinctive feature of TAL is that only auxiliary labels are available in the target data, possibly due to the cost of acquiring the primary labels. This is in contrast to the norm in transfer learning, where primary labels are provided for both source and target tasks. Thus, the key question for TAL is how to best relate the target and source tasks via only auxiliary data in order to produce useful primary target predictions. In the following sections we introduce our proposed TAL models and approaches, which are instantiated and applied to the real-world temporal problem of cold-hardiness prediction. 

\section{TAL Models and Approaches}
\label{sec:approaches}

Motivated by prior work showing the effectiveness of multi-task learning (MTL) for cold-hardiness prediction, our proposed TAL approaches follow a common schema of: 1) learn an MTL model over the source tasks that jointly predicts primary and auxiliary labels, and 2) use the MTL model in different ways, along with the target auxiliary labels, to derive a target model over primary labels. Below we first describe the MTL models considered in this work, followed by our model-selection and model-averaging approaches to using those models for TAL. Finally, we describe the instantiation of TAL for our motivating problem of cold-hardiness prediction. 

\subsection{Multi-Task Models}

\begin{figure}[t]
  \centering
  \includegraphics[width = 0.9\columnwidth]{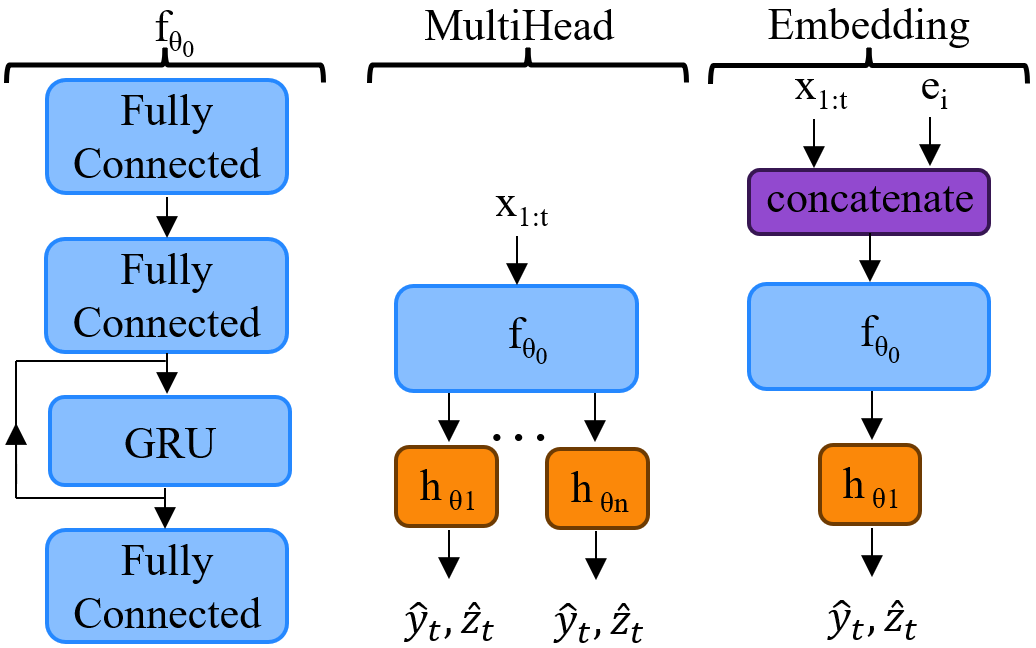}
  \caption{(left) Diagram of the backbone encoder model $f_{\theta_0}$ used for cold-hardiness prediction. Since the inputs for cold-hardiness prediction are time series, this encoder is a recurrent network that encodes any time series into a fixed length vector. (middle) The MultiHead MTL architecture can use any form of backbone encoder to produce primary and auxiliary outputs for each source task. (right) The Embedding MTL architecture concatenates the input with a learned task specific embedding vector and passes this concatenation to encoder and feeds the output to the shared prediction head.
  }
\label{fig:models}
\end{figure}

MTL aims to combine the data from all source tasks in order to arrive at a joint model that can make predictions for all tasks. Below we will denote the model produced by MTL for task $i$ as $M_i$. We focus on two MTL model architectures, which have been successfully applied for cold-hardiness and budbreak prediction in previous work \cite{c:saxena2023,saxena2023b}. Figure \ref{fig:models} depicts these architectures, which are described below.

Both architectures involve a common backbone encoder model $f_{\theta_0}(x)$ with parameters $\theta_0$, which we assume is implemented as a differentiable network function. The role of the encoder is to encode prediction-problem inputs into a fixed-length vector form. For example, for cold-hardiness prediction, each input is a time series of weather data variables and the encoder corresponds to a recurrent GRU network (Figure \ref{fig:models}). The following two MTL model types use this backbone.



\subsubsection{MultiHead Model} This model starts with the encoder and adds a set of prediction heads for each source task, which handle primary and auxiliary predictions. The head for task $i$, denoted $h_{\theta_i}$ is a one-layer fully-connected network with task-specific parameters $\theta_i$, which takes the backbone features as input predicts the primary and auxiliary output, $(y,z)$. Formally, the prediction from $M_i$ given input $x$ is $h_{\theta_i}(f_{\theta_0}(x))$. The backbone parameters $\theta_0$ are shared by all tasks, which was shown to improve predictions over training tasks alone \cite{c:saxena2023}. 


\subsubsection{Embedding Model} Rather than having multiple prediction heads, one per task, the embedding model uses only one prediction head and incorporates the task-specific information at the model input. Specifically, for each source task $i$ we define a dense task embedding vector $e_i$, which is learned during training. We denote by $x^{(i)}$ the concatenation of input data $x$ and embedding vector $e_i$. This model simply provides the backbone encoder with $x^{(i)}$, which encodes both task and input data. The backbone features are then given to a single prediction head with parameters $\theta_1$. Formally, the prediction for $M_i$ given input data $x$ is $h_{\theta_1}(f_{\theta_0}(x^{(i)}))$. 

As shown in \cite{c:saxena2023}, the learned embedding space is able to capture general characteristics of the source tasks and improves per-task predictions compared to single-task training. We aim for an embedding space where the distance between task embeddings relates to the similarity of tasks and where random embedding vectors correspond to meaningful tasks outside of the source task set. In our experiments, we show that for our motivating application, random embedding vectors, which correspond to random new cultivars, do exhibit the qualitative characteristics of grape cold-hardiness and phenology, while differing from the source tasks. Our model-averaging approach, described below, will exploit this property. 


\subsection{TAL via Model Selection}

Since the MTL model can make predictions for multiple tasks, a natural approach to TAL is to select a ``surrogate task" to predict the primary label for the target task. We consider the two following methods for selecting a surrogate task. 


\subsubsection{Best Source} 
This approach assumes there is a source task which produces quality primary label predictions for the target task and can be identified by good auxiliary label predictions. Specifically, given a target task dataset $D_{i^*}$ (containing only auxiliary labels) this method identifies a source task 
$$\hat{i} = \arg\min_{i\in T_s} L(M_i,D_{i^*})$$ with minimal target prediction loss $L(M,D)$. 
We use $M_{\hat{i}}$ to predict primary labels for target task $i^*$.



\subsubsection{Optimized Embedding}
Note that the embedding vector used by the Embedding model is not limited to the learned source task embeddings $\{e_i\}$. That is, any vector from the task embedding space can be treated as specifying a task and used to make predictions. This approach identifies an embedding vector that optimizes the embedding model's target auxiliary predictions and uses that embedding vector to predict target primary labels. Specifically, let $M_e$ denote an instance of an embedding model that uses embedding vector $e$. Given a target auxiliary task $D_{i^*}$ we compute the best embedding vector $$\hat{e} = \arg\min_{e\in E} L(M_e, D_{i^*})$$ using gradient descent, where $E$ is the space of all embedding vectors. $M_{\hat{e}}$ predicts primary labels for the target task.



\subsection{TAL via Model Averaging}

As our experiments show, the model-selection approaches tend to be brittle in our cold-hardiness application and are often not as effective as simple baselines. This brittleness may be due to the relatively sparse information provided by the auxiliary task (phenology in our application), which can lead to many models having similar auxiliary losses yet non-trivial differences in primary labels (LTE). Thus, there is not a strong basis to select a single best model from that set for LTE prediction. Rather than select one model, Model Averaging considers many models and averages them based on their fit to the target auxiliary data. This is similar to the principle of Bayesian Model Averaging, where a posterior over models is used to average over that model space for predictions. 

Formally, we assume that each task is governed by a probability model $\Pr\left(y,z \;|\; x,i\right)$, the probability of the primary and auxiliary labels given the input data and task. Note that the MTL model $M_i$ for each source task is trained to approximate this distribution. We assume the primary and auxiliary labels are conditionally independent given the input $x$ and task $i$ and let $M^y_i$ and $M^z_i$ denote $M_i$'s output of just $y$ or $z$ respectively. 


In the TAL setting, given auxiliary target data $D_{i^*}$, our goal is to compute the probability of the primary target labels, $\Pr\left(y \;|\; x, D_{i^*}\right)$. We define our model averaging approach using the following mixture model over the set of all tasks $T$, which we assume to be finite for notational convenience. 
\begin{small}
\begin{align}
\label{eq:mixture}
\Pr\left(y \;|\; x, D_{i^*}\right) &= \sum_{i \in T} \Pr\left(y\;|\; x,i,D_{i^*}\right)\Pr\left(i\;|x,\;D_{i^*}\right) \nonumber \\
                                    &= \sum_{i \in T} \Pr\left(y\;| x,i\right)\Pr\left(i\;|x,\;D_{i^*}\right) \nonumber \\
                                    &\approx \sum_{i \in T} \Pr\left(y\;| x,i\right)\Pr\left(i\;|D_{i^*}\right)
\end{align} 
\end{small}
The second line follows by noting the conditional independence of the primary and auxiliary labels and that $D_{i^*}$ contains only auxiliary labels. The third line follows from the assumption that task $i$ is approximately independent of the input $x$. This is a reasonable assumption in many applications. If the assumption is strongly violated, the approach 
can be extended to include additional modeling terms for this dependence. 

Equation \ref{eq:mixture} is the basis of our TAL model-averaging approach, which uses the learned MTL model to approximate a weight $w_i = \Pr\left(i\;|\;D_{i^*}\right)$, corresponding to the target task posterior. This is used for a weighted average over individual task models $\Pr\left(y \;|\; x,i\right)$. In order to approximate the task posterior $w_i$ we apply Bayes rule to get $\Pr\left(i\;|\;D_{i^*}\right) \propto \Pr\left(D_{i^*} \;|\; i\right)\Pr\left(i\right)$. If we assume that the prior over tasks is uniform, then we can approximate the posterior by $w_i \approx \Pr\left(D_{i^*} \;|\; i\right)/Z$, where $Z$ is a normalizing constant. If this assumption is violated then any knowledge about non-uniform priors can be easily incorporated into the above process. Note that computing this approximation simply requires computing the log-likelihood of the target auxiliary data under the model for task $i$, which can be done using our learned models $M^z_i$.

\subsubsection{Practical Instantiations}
The general form for our model-averaging TAL approach is to make predictions using
\begin{small}
$$\Pr\left(y \;|\; x, D_{i^*}\right)=\sum_{i \in T} w_i \cdot M^y_i(x),$$
\end{small}
where we use the trained MTL models to calculate the weights. 
We now develop a family of model-averaging instantiations which differ in the way the weights are computed, including selecting non-zero weights. Each instantiation is defined by the following steps: \textbf{1) (Task Selection)} Choose a method to select a set of tasks $\hat{T}\subset T$ that will be assigned non-zero weight, and \textbf{2) (Weight Calculation)} Choose a method to compute a weight value for the each task in $\hat{T}$. Below we describe the choices for these steps that are evaluated in our experiments.


\subsubsection{Task Selection} The first task selection choice is to simply use the set of \emph{source (S) tasks}, i.e. $\hat{T}=T_s$. This choice relies on the source set including one or more tasks that are similar enough to the new target task. To help overcome this potential limitation, when using the embedding MTL model, we can extend the set of tasks by adding randomly generated embeddings. Ideally, the random embeddings correspond to ``fictitious tasks" that are reflective of real tasks and include a wider range of plausible primary label behavior beyond just the source tasks. As our experiments will show, random embeddings for our cold-hardiness prediction problem (corresponding to fictitious cultivars) demonstrate these properties.  

We use two methods for generating random embeddings: \emph{constrained random (CR)} and \emph{random linear combination (LR)}. CR embedding vectors are generated by sampling a value for each embedding feature from a uniform distribution constrained by the maximum and minimum feature value in the source embeddings. Since random embeddings may not be on the manifold of trained source embeddings, there is no guarantee that they will generate realistic predictions. To address this potential issue, LR embeddings are further constrained to be a random convex linear combination of either all or a random subset of the source embeddings. 

\subsubsection{Weight Calculation} We consider multiple approaches for approximating the weight $w_i$ for each task $i\in \hat{T}$. First, we include a baseline method, Uniform, which uses the uniform weighting $w_i=1/|\hat{T}|$. This is the natural model average to use in the absence of any other information about the target task. Our second method attempts to approximate the task posterior given the target data $D_{i^*}$ by an exponentially weighted log-likelihood loss. Specifically, $w_i = exp\left(-\tau\cdot L(M_i,D_{i^*})\right)$, where $L(M_i,D_{i^*})$ is the log-likelihood loss 
and $\tau$ is the temperature parameter, which influences the skewness of the approximated posterior. Finally, we experimentally consider a linear weighting, which simply uses the actual log-likelihood loss for each weight, i.e. $w_i = L(M_i,D_{i^*})$.

\subsection{TAL Instantiation for Cold Hardiness}
\label{sec:tal-for-cold-hardiness}

In our motivating application of cold-hardiness prediction, the different tasks correspond to different grape cultivars. Cold-hardiness prediction is a time-series prediction problem, since the cold hardiness and phenological stage of a grape plant on day $t$ depends on the weather from the start of dormancy to $t$. In particular, each example in the training data has the form $(x_{1:t}, y_t, z_t)$, where $t$ indexes the number of days into the dormancy period starting from a fixed date, $x_{1:t}$ is the weather data time series from the start of dormancy to day $t$, $y_t$ (primary label) is the LTE cold-hardiness value on day $t$, and $z_t$ (auxiliary label) indicates the phenological stage on day $t$. Since the grape data only has cold-hardiness measurements on days when scientists visit the field, we set $y_t = N/A$ on days when the cold-hardiness is not available. Importantly, learned cold-hardiness models are still able to make predictions for any day given the historical weather sequence $x_{1:t}$.

To adapt TAL for our specific problem, we must select a backbone encoder model which can handle time series data. Following previous work \cite{c:saxena2023} on cold hardiness prediction we use a recurrent neural network for encoding the input sequence of weather data $x_{1:t}$. Figure \ref{fig:models} illustrates this encoder, which is composed of two fully connected layers, followed by a gated recurrent unit (GRU) layer, followed by another fully-connected layer. The result is a vector encoding of any sequence $x_{1:t}$, which is then used as the basis for the multi-head and embedding MTL models.

\section{Experimental Setup}\label{sec:exp-setup}

\subsection{Datasets and Training}

\begin{table}[ht]
  \caption{Number of seasons and LTE samples per cultivar. LTE Seasons indicate the number of seasons with at least one LTE sample, Pheno. Seasons the number with all four phenology dates (first swell, full swell, budbreak, first leaf), and Mutual Seasons the number with both. LTE Samples are the total number of LTE samples recorded in all mutual seasons.}
  \centering
  \begin{tabular}{lllll}
    \toprule
    ~        & LTE     & Pheno.  & Mutual  & LTE \\
    Cultivar & Seasons & Seasons & Seasons & Samples \\
    \midrule
    Barbera            & 14  & 7     & 7    & 84          \\
    Cabernet Sauvignon & 32  & 16    & 16   & 465         \\
    Chardonnay         & 25  & 18    & 15   & 450         \\
    Chenin Blanc       & 17  & 17    & 9    & 109         \\
    Grenache           & 14  & 16    & 8    & 97          \\
    Malbec             & 17  & 16    & 8    & 140         \\
    Merlot             & 25  & 20    & 17   & 489         \\
    Mourvedre          & 12  & 7     & 5    & 69          \\
    Nebbiolo           & 14  & 7     & 7    & 85          \\
    Pinot Gris         & 17  & 16    & 9    & 111         \\
    Riesling           & 32  & 18    & 18   & 408         \\
    Sangiovese         & 15  & 7     & 7    & 86          \\
    Sauvignon Blanc    & 12  & 7     & 7    & 87          \\
    Semillon           & 13  & 16    & 7    & 119         \\
    Syrah              & 22  & 4     & 4    & 106         \\
    Viognier           & 17  & 8     & 5    & 62          \\
    Zinfandel          & 14  & 17    & 8    & 94          \\
    \bottomrule
  \end{tabular}
  \label{tbl:data-count}
\end{table}

This work uses grape cold-hardiness and phenological data collected from 1988 to 2022 at the WSU Irrigated Agriculture Research and Extension Center in Prosser, WA. Cane samples containing dormant buds were collected and analyzed to determine the ground truth LTE values. Varying numbers of phenological events were observed during each season for each cultivar. Daily weather data from the closest weather station is from the AgWeatherNet API \cite{c:agnet}.

The dataset is constructed using dormant season data from September 7th to May 15th. Each season must have less than 10\% of the weather data missing, at least one LTE sample, and occurrence dates for four phenological events: first swell, full swell, budbreak, and first leaf
, which are the most prevalent in the dataset. The number of satisfactory seasons varies per cultivar, ranging from 4 to 18 seasons each, with an average of 19.5 samples per season. Table \ref{tbl:data-count} lists the per-cultivar data totals of the dataset. 
Missing weather data is filled with linear interpolation and missing LTE values are masked out during training. The 12 features are min/max/avg air temperature, min/max/avg relative humidity, min/max/avg dew point, precipitation, and max/avg wind speed.

Training and test sets are created by removing two seasons per cultivar for the test set and using the rest for training. The presented results are averages over three different training-test trials. The models are trained with the Binary Cross-Entropy of the phenology predictions at each time step and the Mean Square Error (MSE) of the $LTE_{10}$, $LTE_{50}$, and $LTE_{90}$ predictions at time steps with ground-truth LTE available. The performance is evaluated using the Root Mean Square Error (RMSE) of the $LTE_{50}$ prediction. For training, the Adam optimizer \cite{c:kingma2017} is used with a learning rate of 0.001. Complete training details and network architecture are given in Appendix \ref{sec:implementation-appendix}.
 
Prior work trained separate MTL models for predicting LTE \cite{c:saxena2023} and phenology \cite{saxena2023b}. Instead, we train MTL models that predict both LTE and phenology by including 7 prediction heads for each cultivar (3 for LTE and 4 for phenology). It is possible that combined training may hurt the accuracy of our primary LTE labels, which would reduce the potential impact of TAL. In order to test this possibility, we compare the RMSE of LTE predictions for the Embedding (1.39, 1.40) and MultiHead (1.31, 1.28) models trained with just the LTE labels and with both the LTE and phenology labels, respectively. We see that for both models, including auxiliary information in training has no significant impact on the average primary prediction performance. For the remainder of the paper, all models are trained to predict both primary and auxiliary labels. 

\subsection{Evaluation Protocol and Comparisons}
We evaluate the performance of our proposed TAL methods on the grape 
datasets. For each cultivar $i \in C$, we train a base MTL model $M^i$ on the set of source cultivars $C\backslash\{i\}$. This model allows us to simulate transfer to cultivar $i$ where all other cultivars are source tasks. The same $M^i$ is used by all TAL experiments that are transferring to $i$. All results provide the RMSE for LTE prediction on each cultivar and the average of all cultivars.

As described in Section \ref{sec:related}, we have been unable to find prior transfer learning work that is directly or easily modified to be applicable to our TAL framework. Thus, our experiments are focused on comparing the different approaches from Section \ref{sec:approaches}, including comparisons to baselines that are strong in that they significantly outperform some of the proposed approaches. Similarly, there are no prior transfer-learning benchmarks that are easily adapted to evaluating TAL performance.  One point of future work is to develop additional TAL benchmarks, for example, based on appropriate datasets from agriculture and health. In this work, we focus on a careful evaluation within our real-world grape cold-hardiness dataset, which has immediate practical utility to farmers. 

\section{Experimental Results}

\begin{sidewaystable*}[htp!]
  \caption{Comparison of RMSE of LTE prediction for the different TAL approaches for both the Embedding and MultiHead models. The column labels denote: 1) the type of model used
  , 2) the type of TAL scheme
  , 3) the specific TAL scheme and the model set
  . Note that Pheno. and LTE on the Optimization columns refer to the loss used in the optimization. * Oracle method that uses LTE training data for target tasks, which is intended to give an optimistic bound on performance.}
  \small
  \centering
  \begin{tabular}{llllllll|lll}
    \toprule
    ~ & \multicolumn{7}{c|}{Embedding Model} & \multicolumn{3}{c}{MultiHead Model} \\
    \cmidrule(r){2-11}
    ~ & \multicolumn{4}{c}{Model Averaging} & \multicolumn{3}{c|}{Selection} & \multicolumn{2}{c}{Model Averaging} & Selection \\
    \cmidrule(r){2-5}
    \cmidrule(r){6-10}
    \cmidrule(r){11-11}
    
    ~ & Uniform & Weighted & Uniform & Weighted & Best & Pheno. & LTE & Uniform & Weighted & Best \\
    Cultivar & (S) & (S) & (S+CR) & (S+CR) & Source & Optim. & Optim.* & (S) & (S) & Source \\
    \midrule
    Barbera         & 2.10 & 1.56 & 1.51 & 1.48 & 1.42 & 1.48 & 1.40 & 1.56 & 1.55 & 1.51 \\
    Cabernet S.  & 1.21 & 1.33 & 1.33 & 1.33 & 1.37 & 1.52 & 1.28 & 1.70 & 1.72 & 1.66 \\
    Chardonnay      & 1.77 & 1.39 & 1.86 & 1.49 & 2.07 & 1.39 & 1.41 & 1.58 & 1.48 & 3.40 \\
    Chenin Blanc    & 1.19 & 1.35 & 1.48 & 1.57 & 2.79 & 1.36 & 1.00 & 1.30 & 1.30 & 1.56\\
    Grenache        & 1.97 & 2.24 & 2.31 & 2.39 & 2.85 & 2.11 & 1.56 & 2.40 & 2.35 & 1.64\\
    Malbec          & 1.22 & 1.14 & 1.13 & 1.07 & 1.51 & 1.16 & 1.33 & 1.13 & 1.12 & 1.63 \\ 
    Merlot          & 1.59 & 1.21 & 1.23 & 1.27 & 2.16 & 1.70 & 1.27 & 1.38 & 1.40 & 2.44 \\
    Mourvedre       & 1.82 & 1.76 & 1.81 & 1.34 & 1.34 & 1.36 & 1.08 & 2.03 & 1.90 & 1.38 \\
    Nebbiolo        & 1.54 & 1.39 & 1.44 & 1.40 & 1.53 & 1.56 & 1.23 & 1.36 & 1.36 & 1.53\\
    Pinot Gris      & 1.40 & 1.52 & 1.35 & 1.39 & 2.25 & 1.66 & 1.36 & 1.37 & 1.38 & 2.44 \\
    Riesling        & 2.02 & 1.71 & 1.80 & 1.65 & 2.52 & 1.90 & 1.39 & 2.21 & 2.17 & 3.56 \\
    Sangiovese      & 2.26 & 1.95 & 1.51 & 1.48 & 2.03 & 2.14 & 1.51 & 2.23 & 2.24 & 1.46 \\
    Sauv. Blanc & 1.16 & 1.02 & 1.03 & 0.98 & 1.45 & 1.26 & 1.09 & 1.13 & 1.12 & 1.49 \\
    Semillon        & 2.47 & 2.00 & 2.13 & 2.06 & 1.45 & 1.58 & 1.47 & 2.11 & 2.08 & 1.54 \\
    Syrah           & 1.17 & 1.32 & 1.24 & 1.32 & 1.39 & 2.06 & 1.51 & 1.62 & 1.68 & 1.81 \\
    Viognier        & 1.23 & 1.42 & 1.52 & 1.49 & 1.48 & 1.14 & 1.27 & 1.30 & 1.33 & 2.57 \\
    Zinfandel       & 1.25 & 1.20 & 1.16 & 1.17 & 1.50 & 2.09 & 1.54 & 1.51 & 1.47 & 1.57 \\
    \midrule
    Mean            & 1.61 & 1.50 & 1.52 & 1.46 & 1.83 & 1.62 & 1.33 & 1.64 & 1.63 & 1.95 \\ 
    \bottomrule
  \end{tabular}
  \label{tbl:all}
\end{sidewaystable*}

We consider various methods and summarize the results in Table \ref{tbl:all}. For TAL methods with free parameters (e.g. the type and number of random embeddings), the table uses parameter settings based on initial exploratory investigations. Later experiments in this section explore parameter sensitivity.

\subsection{Overall Benefit of TAL}
Results for all of our TAL methods are shown in Table \ref{tbl:all}. Overall we see that for both the Embedding and MultiHead model types, all TAL Model Averaging methods perform better than or are competitive with the baseline method Uniform (S). Note that for the Embedding model, just adding weighting (Weighted (S)) or just adding more models (Uniform (S+CR)) improves performance over the baseline. Combining these two methods (Weighted (S+CR)) amplifies this effect, performing the best out of all methods. This best method does not achieve the oracle performance of LTE Optimization or supervised training (Section \ref{sec:exp-setup}), which are able to use LTE information for the target cultivar. However, TAL has significantly closed the gap between oracle methods and the baseline by exploiting target auxiliary information. This improvement over the natural baseline is especially important given that there are no other prior methods that could handle this specific problem setting.

The improvement of Uniform S+CR is particularly interesting since it is not taking advantage of the auxiliary target information. One hypothesis for this improvement is that by including random embeddings that capture a diverse, but plausible, range of behavior, the uniform model average gets closer to the mean LTE model over all potential target cultivars. Among all fixed models, such a mean model would minimize the expected error over target cultivars. 

In contrast, TAL Model Selection does not see the same performance improvements. These methods, surprisingly, perform worse than Uniform (S) for both base model types. This suggests that choosing a single model based on the auxiliary labels is brittle compared to the softer approach of weighting models, possibly due to overfitting. With enough cultivar models and some random noise in the data, there is a reasonable probability that a sub-optimal cultivar is chosen.

\subsection{MultiHead vs. Embedding Model}
A key observation from Table \ref{tbl:all} is that, on average, all TAL methods using the Embedding model outperform those using the MultiHead model. 
This is interesting since Section \ref{sec:exp-setup} shows the opposite relative performance for standard MTL supervised training. Two possibilities for this performance difference are: 1) the LTE predictions for source cultivars from the Embedding model transfer better on average to new cultivars, and 2) the weighting over source cultivars produced by the Embedding model better reflects the source-target relationship. 


\begin{table}[t!]
  \caption{Comparison of RMSE of LTE predictions averaged over all cultivars using varying weighted model averaging methods on the Embedding (Emb) and MultiHead (MH) models. Both models are evaluated using weighting schemes from both their own and the other model's prediction losses.}
  \centering
  \begin{tabular}{lllll}
    \toprule
    Model,Weighting & Lin & Ex-5 & Ex-10 & Ex-20 \\
    \midrule
    MH, Emb (Mean)  & 1.65 & 1.65 & 1.66  & 1.69  \\
    MH, MH (Mean)   & 1.62 & 1.63 & 1.69 & 1.78  \\
    Emb, MH (Mean)  & 1.52 & 1.53  & 1.53 & 1.55 \\
    Emb, Emb (Mean) & 1.50 & 1.50 & 1.50 & 1.53 \\
    \bottomrule
  \end{tabular}
  \label{tbl:mixed-weighting}
\end{table}

To explore these possibilities we ran TAL experiments where we averaged one model's LTE predictions using the other model's cultivar weights. Table \ref{tbl:mixed-weighting} shows the results for 4 different weighting schemes. We found that for all 4 schemes, the Embedding model using the MultiHead weights significantly outperformed the MultiHead model using the weights from either model. This gives strong evidence that the Embedding model's LTE predictions are better suited to transfer to new unseen cultivars. That is, the cultivar-specific LTE predictions from the Embedding model are closer to a mean LTE model. In contrast, the MultiHead model produces LTE predictions that may capture finer details of each source cultivar, which yields stronger supervised performance but decreases expected performance on new cultivars. 

We also found that the Embedding model always performed better using its own weights compared to using the MultiHead model weights. Rather, only half of the time did the MultiHead model perform better with its own weights. This suggests that the Embedding model can better capture the source and target cultivar similarity via the auxiliary labels. A potential reason for this is that the Embedding model's architecture includes the cultivar information at the input of the network, forcing it to create a more direct connection between phenology and LTE than the MultiHead model, which distinguishes between cultivars at the output via separate heads.  

\subsection{Impact of Embedding Choices}
Here we try different embedding sets for model averaging. For LR, we consider two types, where each embedding is a combination of: 1) all the source embedding (LR-17), and 2) a random subset of three source embeddings (LR-3). 

We calculate the mean of 17 cultivars' RMSE for Uniform weighting LTE predictions, using the number of embeddings that performed best for each method (68 for CR and 17 for LR), noting that comparative conclusions are the same for other numbers of embeddings. We find that 
CR (1.53) and S+CR (1.52) perform the best and outperform the baseline method (S, 1.61). In contrast, S+LR-3 (1.72) and S+LR-17 (1.95) are worse than the baseline and 
their performance worsens as more embeddings are added (not shown).

\begin{figure}[t!]
  \centering
  \includegraphics[width = 0.85\columnwidth]{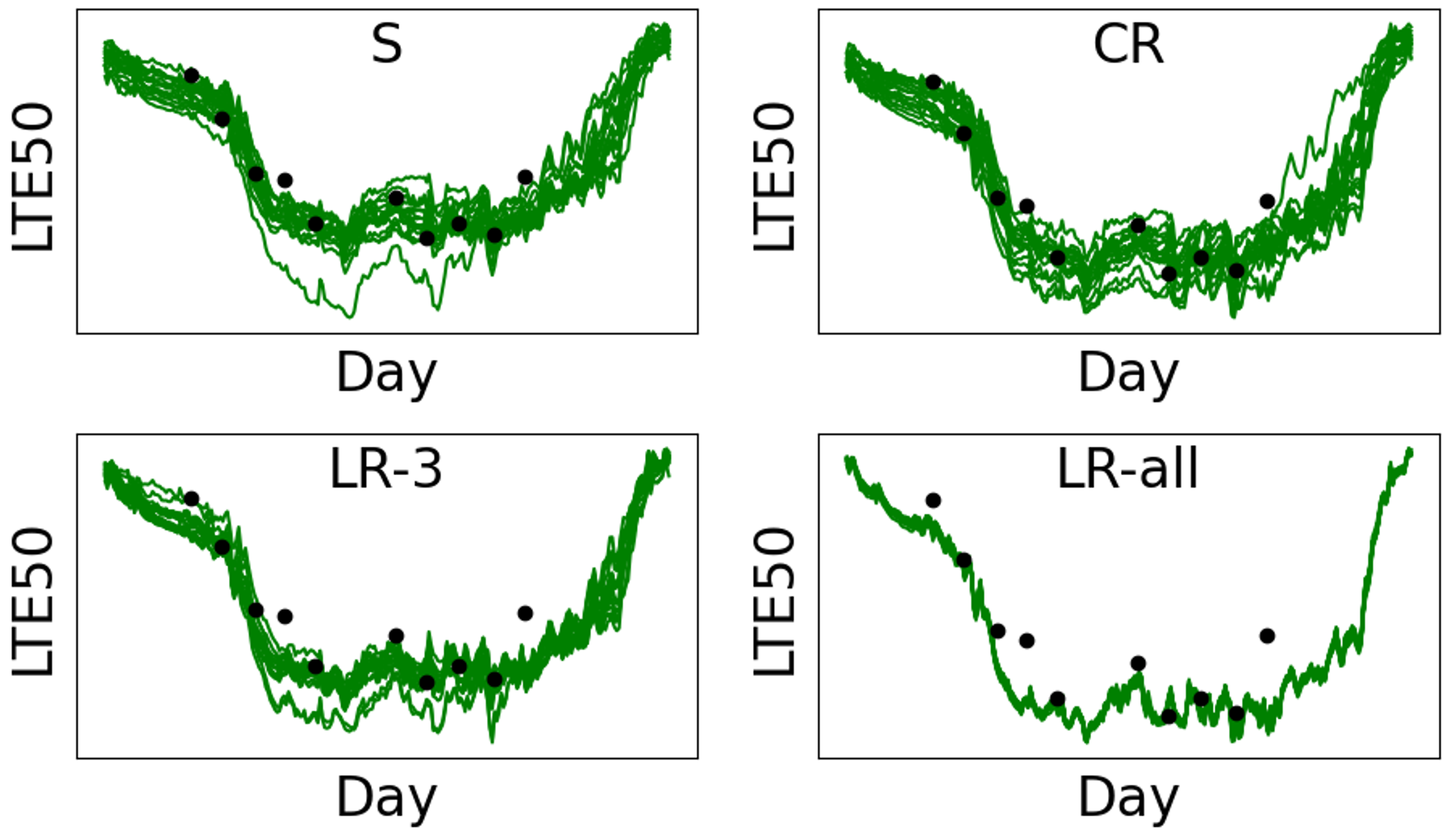}
  \caption{Comparison of different embedding sets (S, CR, LR-3 and LR-17). The green lines show the LTE predictions for the 2020 season using embeddings from a model trained with Barbera as the target task and all others as the source tasks. The black dots indicate the Barbera ground truth LTE values.}
\label{fig:emb-space}
\end{figure}

To investigate these performance differences, 
Figure \ref{fig:emb-space} plots the LTE predictions of individual embeddings for the 2020 season using the Barbera-transfer model (i.e. $M^{i=\mbox{Barbera}}$). First, we observe that both the CR and LR approaches produce curves that are qualitatively plausible LTE responses. This is particularly interesting for CR, since its embeddings are not constrained to any particular manifold. Second, we observe that S and CR embeddings have a much wider range of output behavior compared to LR. As a result S and CR embeddings more fully encompass the ground truth LTE targets 
than LR which often does not include the ground truth. This explains the poor performance of Uniform for LR, since the model average puts too much influence on the inaccurate embeddings.  

\subsection{Impact of Embedding Set Size}

\begin{table}[t!]
  \caption{Comparison of mean cultivar RMSE of LTE predictions from Uniform and Weighted (exp, $\tau=10$) Embedding model (S+CR) with varying numbers of CR embeddings.}
  \centering
  \begin{tabular}{llllll}
    \toprule
    Num. CR Embeddings & 17 & 34 & 68 & 136 & 272 \\
    \midrule
    Uniform (Mean)& 1.58 & 1.57 & 1.52 & 1.56 & 1.55 \\
    Weighted (Mean) & 1.53 & 1.51 & 1.46 & 1.51 & 1.51 \\
    \bottomrule
  \end{tabular}
  \label{tbl:num-embs}
\end{table}

Table \ref{tbl:num-embs} shows the influence of varying the number of CR embeddings in S+CR for Uniform and exponential weighting. Results for CR are qualitatively similar (not shown). As expected, weighting outperforms Uniform for all cases, showing it can leverage the auxiliary target labels. For both Uniform and exponential weighting, as the number of embeddings increases, the performance improves until reaching 68 embeddings and then starts to level off and degrade. A possible explanation is that as the number of random embeddings grows, the model average approaches a canonical model that captures average behavior, yielding improved expected error on new cultivars. However, as the number of models grows there is also more of a chance that outlier embeddings are produced, which may produce unusual LTE behavior that can hurt the overall model average.


\subsection{Weighting Scheme Analysis}

\begin{table}[t!]
  \caption{Comparison of mean cultivar RMSE of LTE predictions for embedding sets with varying weighted model averaging methods. Ex-n is exponential weighting with $\tau=n$.}
  \centering
  \begin{tabular}{llllll}
    \toprule
    Embedding Type & Lin & Ex-5 & Ex-10 & Ex-20 & Ex-50 \\
    \midrule
    S (Mean)& 1.50 & 1.50 & 1.50 & 1.53 & 1.59 \\
    CR (Mean) & 1.50 & 1.48 & 1.47 & 1.48 & 1.49 \\
    S+CR (Mean) & 1.49 & 1.47 & 1.46 & 1.47 & 1.48 \\
    \bottomrule
  \end{tabular}
  \label{tbl:wsa-s}
\end{table}

To investigate the influence of weighting schemes, Table \ref{tbl:wsa-s} shows the average performance of several embedding sets for model averaging with linear and exponential weighting. We see that all options outperform the baseline (Uniform S), indicating that all choices can leverage the auxiliary target labels. When using just set S, linear weighting and exponential with smaller $\tau$ perform similarly and performance gets worse for exponential with larger $\tau$. The decreasing performance can be explained by the fact that large $\tau$ values results in sparse weightings, which makes model weighting behave similarly to model selection over S, which was shown to perform poorly (Table \ref{tbl:all}).  

Results for CR and S+CR show a small improvement in performance for exponential with smaller $\tau$, with the optimum at $\tau=10$, and then a slight decrease in performance. This shows that tuning the exponential parameter for the more diverse set of embeddings has value, but is not overly critical. Note that CR and S+CR outperform S, showing that the methods can exploit the extra embeddings. 


\section{Conclusion}
We considered how to leverage easy-to-collect phenological crop data to support the prediction of hard-to-collect cold-hardiness data. For this purpose, we introduced a novel transfer learning framework where transfer to a target cultivar is enabled via auxiliary labels (phenology). We introduced several approaches to this problem and demonstrated that model averaging using an embedding-style multi-task neural network was most effective. This work proposes a preliminary method for an apparently under-explored problem setting. We expect that there is still room to develop improved algorithms for this transfer setting and that many other problems in agriculture and other domains can also benefit from this framework. 



\section*{Declarations}

\bmhead{Data availability}
Data are available upon reasonable request.



\backmatter

\begin{appendices}

\section{Implementation Details}
\label{sec:implementation-appendix}

The backbone model used in this work is a neural network with four layers. The first layer has 1024 fully connected nodes that output to a ReLU layer. The second layer has 2048 fully connected nodes that output to a ReLU layer. The third layer is a Gated Recurrent Unit (GRU) with a single layer of 2048 nodes. The fourth layer has 1024 fully connected nodes that output to a ReLU layer. The output of the backbone model is then fed to the appropriate prediction heads. Each set of prediction heads has seven fully connected single nodes, three for predicting LTE and four for predicting phenology. The MultiHead model has a set of prediction heads for each cultivar, whereas the Embedding model uses the same set of prediction heads for all cultivars. The Embedding model also has an additional 12-dimensional embedding vector which is concatenated with each time step of the input weather data before being fed into the backbone model.

For training, the Adam optimizer \cite{c:kingma2017} is used with a learning rate of 0.001. The batch size is 12, shuffled, and the model is trained for 400 epochs. The models are trained with the Binary Cross-Entropy of the phenology predictions at each time step and the Mean Square Error (MSE) of the $LTE_{10}$, $LTE_{50}$, and $LTE_{90}$ predictions at time steps with ground-truth LTE available. The performance is evaluated using the Root Mean Square Error (RMSE) of the $LTE_{50}$ prediction.

\end{appendices}

\bibliography{sn-bibliography}

\end{document}